\documentclass[10pt,letterpaper]{article}

\usepackage{cogsci}
\usepackage{xcolor}
\usepackage{siunitx}
\usepackage{url}
\cogscifinalcopy 

\usepackage{pslatex}
\usepackage{apacite}
\usepackage{float} 

\usepackage{graphicx}

\usepackage{xcolor}

\title{The Diversity of Argument-Making in the Wild: from Assumptions and Definitions to Causation and Anecdote in Reddit's ``Change My View''}
 
\author{{\large \bf Robin W. Na (robinna@mit.edu)}  \\
  Sloan School of Management, Massachusetts Institute of Technology, Cambridge, MA 02142 USA
  \AND {\large \bf Simon DeDeo (sdedeo@andrew.cmu.edu)} \\
  Department of Social and Decision Sciences, Carnegie Mellon University, Pittsburgh, PA 15213 USA \\ Santa Fe Institute, Santa Fe, NM 87501 USA}

\begin{document}

\maketitle

\begin{abstract}
What kinds of arguments do people make, and what effect do they have on others? Normative constraints on argument-making are as old as philosophy itself, but little is known about the diversity of arguments made in practice. We use NLP tools to extract patterns of argument-making from the Reddit site ``Change My View'' (r/CMV). This reveals six distinct argument patterns: not just the familiar deductive and inductive forms, but also arguments about definitions, relevance, possibility and cause, and personal experience. Data from r/CMV also reveal differences in efficacy: personal experience and, to a lesser extent, arguments about causation and examples, are most likely to shift a person's view, while arguments about relevance are the least. Finally, our methods reveal a gradient of argument-making preferences among users: a two-axis model, of ``personal--impersonal'' and ``concrete--abstract'', can account for nearly 80\% of the strategy variance between individuals.

\end{abstract}


\noindent People are not content to simply hold beliefs: they also try to persuade others to hold them as well. While there are many ways to do this, one of the most celebrated forms of persuasion is argument-making. The giving of public reasons to adopt beliefs is a core feature of human culture~\cite{HAHN2020363}; indeed, under the ``argument theory'' of \citeA{mercier2011humans}, it is the origin of deliberative reason itself.

Perhaps because of this, normative taxonomies of argument-making are at the heart of systematic philosophy~\cite{mike}. Aristotle's \emph{Prior Analytics} presented a scheme for different ways in which the truth of a conclusion can be established through deductive argument. Much later, \citeA{bayes} introduced a formal framework for inductive reasoning, on the basis of likelihood, rather than deductive certainty, and the ``deductive vs. inductive'' distinction persists to the present day. In the modern era, cognitive scientists have focused on a variety of ways in which our explanatory values might guide us to prefer one belief over another~\cite{tania, simon, igor}. 

Deductive, inductive, or more broadly explanatory considerations do not exhaust the nature of arguments, however. One reason is that argument-making is often a matter of argument \emph{criticism}. If I expose the faulty assumptions in someone's argument, question the relevance of a point, or reveal an equivocation, for example, I may go a long way towards bringing them towards the opposite belief. This is to be expected from a Bayesian point of view: by reducing someone's confidence in an argument for X, I may increase their confidence in the alternative I believe to be the case---particularly when the argument in question was decisive for them in overcoming what were previously shared priors.

Another complication is the fundamentally intersubjective nature of argument-making~\cite{knauff2021handbook}. The reasons that might convince me to hold a view are not necessarily the same kinds of reasons that I could use to convince others. The priors I have, the assumptions that go into my categories, or the level of logical rigor or causal precision I demand may well differ from my interlocutor, and a full taxonomy of argument-making will naturally take into account considerations similar to those found in the (linguistic) pragmatics of cooperation~\cite{grice}. A successful argument, for example, might hinge on revealing a hidden, but unshared, assumption, explaining why something is actually irrelevant, or on clarifying what was previously obscure between us.

Our goal in this work is two-fold. First: what are the broad patterns in the arguments people actually make? While popular accounts partition argument-making into a ``deductive'' and ``inductive'' type, it is clear that when we argue we do more than discuss probabilities and premises. Second: how, and in what contexts, do these patterns of argument actually work to persuade? Arguments we accept may be quite different from the ones we ought to.

The availability of large corpora provides a new way to answer these questions. Rather than pre-determine the possible kinds of arguments that are being made, we present a method for discovering, in an unsupervised fashion, implicit patterns of argument-making. We do so using an unusual dataset from the Reddit site ``Change My View'', a highly-curated system where users engage in good-faith attempts to change each other's points of view, and rate the outcomes. The volume of text allows us to surface patterns of co-occurring ``argument fragments''---single words such as ``prove'', for example, and bigrams such as ``distinction between''---that can signal different approaches to persuasion. Our use of bigrams, in particular, allows us to go beyond traditional tools, like the LIWC, to capture the contextual ways in which (for example) the phrases ``perfectly possible'' and ``entirely possible'' may signal distinct epistemic strategies.

While our work is related to an active area of research in NLP known as ``argument mining''~\cite{argument_mining}, our goals are very different. Argument mining aims to map a text into a structured form, such as a rhetorical schema or an Aristotelian syllogism, and to tag that text for both role (\emph{e.g.}, ``assertion of a causal relationship'') and content (what the cause and effect actually are). We, by contrast, are interested in the patterns implicit in actual use---whether or not they can map to a pre-existing normative structure.


\section{Methods}

\subsection{Corpora: r/ChangeMyView and LessWrong}

Our main corpus is from the Reddit site r/ChangeMyView (r/CMV; \url{https://www.reddit.com/r/changemyview/}). A poster on r/CMV (the ``original poster'', or OP) presents a view that they hold to the rest of the group; the post begins a discussion where other participants present arguments against that point of view, with the goal of changing the OP's mind. The site is actively, and quite strictly, moderated, and provides a well-curated collection of arguments and counter-arguments about a variety of questions, from political policy to moral and ethical questions, and to abstract questions such as the existence of God. The ``good faith'' discussions of r/CMV have proved fertile ground for understanding persuasion; they have been used in studies of evidence-giving~\cite{priniski2018attitude}, language alignment and turn-taking~\cite{entry_order}, and as raw material for machine-learning prediction tasks (\emph{e.g.}, \citeA{Hidey_McKeown_2018}).

As with many online systems, readers can upvote (or downvote) comments, but in addition, users are encouraged, where appropriate, to tag a comment with a ``delta'', to indicate that the comment changed their point of view.

A recent post on r/CMV, for example, began ``massive companies and local businesses should not be treated the same under the law'', and garnered over a hundred responses in the following six hours.\footnote{See \url{https://bit.ly/cmv_example}} The OP tagged three replies with a ``delta'', indicating that the reply changed their point of view, including a reply that suggested it would be difficult to come up with clear definitions of large vs. small businesses, a reply that large vs. small business were already being treated differently, and a reply that suggested the fix would likely harm, rather than help, the people the OP had in mind. Our final corpus contains 100,170 ``posts'' (\emph{i.e.}, an original post stating a view), and a total of 5,833,572 replies and counter replies to the view; roughly 1\% of the replies are tagged as having changed someone's point of view. 

We use r/CMV to construct our argument patterns. To study how well our results extrapolate to other communities, we also use data from the discussion site ``LessWrong'' (\url{http://lesswrong.org}). LessWrong is a site associated with the ``rationalist'' community, where users make arguments about questions such as the dangers of artificial intelligence and the relative merits of different economic and political systems. LessWrong does not have r/CMV's delta-tagging system, but it does have an upvote/downvote mechanism that allows us to track how the arguments are perceived by other users. Our LessWrong data contains 25,841 posts, and 708,807 replies.


\subsection{Argument Extraction with Linkage Networks}

At heart, argument-making is associated with adjusting degrees of belief. To identify argument patterns in an unsupervised fashion, we thus begin with a seed set of words from the widely-used LIWC collection~\cite{pennebaker1999linguistic}: in particular, those from the wordlists ``Tentative'' (\emph{e.g.}, ``likely,'' ``vaguely'') and ``Certain'' (\emph{e.g.}, ``surely,'' ``clearly''). Because these seed words can convey different meanings depending on their adjacent words, we use standard methods~\cite{mikolov-bigrams} on the r/CMV corpus to locate bigrams containing these terms. Our seed list then contains not only a word like ``true'', but also combinations such as ``factually true'', ``necessarily true'', and ``holds true'' that may further signal different styles of approach.

With this seed set in hand, we then use an information-theoretic method to find larger patterns of co-occurrence. For each pair of words $w_i$ and $w_j$ in this collection, we use the r/CMV data to measure the ``linkage'', or pointwise mutual information, between them,
\begin{equation}
L_{ij} = \log_2{\frac{P(w_i,w_j)}{P(w_i)P(w_j)}}
\end{equation}
where $P(w_i)$ is the probability of drawing word $w_i$ from a random document (r/CMV post) which in our case filtered to only include seed words. $P(w_i,w_j)$ is the probability of drawing $w_i$ and then $w_j$ from a random document, which can be estimated as
\begin{equation}
P(w_i, w_j) = \frac{1}{|\mathcal{D}|}\sum_{d\in\mathcal{D}}\frac{N(w_i, d)N(w_j,d)}{N(d)^2},
\end{equation}
where $N(w_i,d)$ is the number of times word $w_i$ appears in document $d$, and $N(d)$ is the total number of words in document $d$. The linkage $L_{ij}$ measures the extent to which the use of one word predicts the use of another. 

With argument fragments represented as nodes in our linkage network, we then use Louvain clustering to detect communities of highly interlinked argument fragments based on the criteria of highest modularity~\cite{louvain-algorithm}. Unlike in topic modeling, the number of clusters is free to vary and not set ahead of time.

The clusters define the basic patterns of co-occurrence in argument-related words. We then augment our network with candidate words and bigrams from an additional category in LIWC named ``cognitive processes", and from highly-weighted words from an argument-related topic detected through topic modeling of another subreddit, r/TheRedPill (Topic 2 in \citeA{chloe}).

To do this, we measure the linkage between each new candidate fragment and the clusters found in the first step,
\begin{equation} \label{linkage_cluster}
\hat{L}_{kj} = \log_2{\frac{P(C_k,w_j)}{P(C_k)P(w_j)}} = \log_2{\frac{P(C_k | w_j)}{P(C_k)}}
\end{equation}
where $P(C_k)$ is the probability of drawing a word from cluster $C_k$ from a random document and $P(C_k|w_j)$ is the probability of drawing a word from cluster $C_k$ in random documents conditioned upon containing word $w_j$. This can be estimated as
\begin{equation}
P(C_k|w_j) = \frac{1}{Z} \sum_{d\in\mathcal{D}_j}\frac{N(C_k, d)N(w_j,d)}{N(d)}
\end{equation}
where $\mathcal{D}_j$ is a set of documents including word $w_j$, $N(C_k, d)$ is the number of times that a word in $C_k$ appears in document $d$. $N(w_j,d)$ is multiplied to use the information about the frequency of word $w_j$ appearing in document $d$. $Z$ is the normalizing constant for probability.

The new $\hat{L}_{kj}$ represents a bipartite network between argument fragments and clusters. To maximize the modularity of this bipartite network, we first maximize the modularity of the network which only includes seed fragments---this involves a few minor changes in the original cluster memberships. Then, for each new candidate fragment, we admit the candidate only if adding the link results in an increase of modularity, assigning it to the cluster $k$ with maximum $\hat{L}_{kj}$.


\subsection{Topic Modeling of Semantics}

In both r/CMV and LessWrong, we expect that argument patterns will be more or less common---and more or less successful--depending upon the subject matter. A discussion about abortion rights, for example, might hinge on an argument over the definition of ``life'', while a discussion about foreign policy might involve more inductive concerns about the likelihood of success. This is both interesting in its own right, and a source of potentially spurious correlation. It may be the case (for example) that ``definitional'' arguments are more common in abortion debates, but it also may be the case that the underlying moral commitments of the participants make it less likely for people to change their mind compared to less hot-button topics.

In order to separate posts into different semantic categories, we build a topic model~\cite{McCallumMALLET} of the texts, after removing argument words; because we are also interested in studying the intersubjective aspect of argument-making, we remove pronouns from this list as well. The resulting topics allow us to classify posts into combinations of different semantic themes; in nearly every case, one theme is dominant, and this enables us to separate out posts to look at the relationship between argument pattern (``pragmatics'') and subject matter (``semantics''). In addition, the topic model allows us to identify, and automatically remove, non-argument topics such as ``meta'' discussion (discussion of the rules of the site, complaints to moderators, and so on) and, in the case of LessWrong, a small component of Harry Potter fan fiction.

\section{Results}

We report three main results: (1) the nature of the argument patterns discovered by the linkage network method, (2) the relative effectiveness of these patterns in changing a person's view, and (3) the diversity of argument-making preferences at the level of the individual.

\subsection{Argument Patterns}

The linkage network method ends up allocating a total of 1,506 unigrams and bigrams into six distinct clusters or argument patterns. These six patterns are shown in Table~\ref{linkage_table}. We list sample bigrams from each cluster (these are more interpretable than the unigrams), along with provisional names that characterize the kinds of arguments the bigrams tend to appear in. To help support our provisional naming choices, Table~\ref{examples} provides examples of comments from r/CMV that are heavily-weighted on each pattern.


Two of the patterns we find, ``deduction and certainty'' (P3), and ``induction and probability'' (P5), correspond to classic distinctions often made between, on the one hand, arguments based on certainties and definitive truths, amenable to proof and logical deduction, and, on the other, arguments based on relative likelihoods, average cases, and in conditions of potential uncertainty. A third pattern, ``causation and examples'' (P4), includes both causal vocabulary and words associated with example-giving; in both cases, the connection is to modal notions of possibility.

Two of the patterns have a strongly intersubjective aspect, resonating with Grice's cooperative maxims~\cite{grice}. ``Relevance and presumption'' (P1) includes arguments where the writer draws attention to, and critiques, the relevance of a point made in an an argument; this includes both direct criticism of assumptions, as well as implicit responses (\emph{e.g.}, ``never said'', as in ``I never said that X [but you assumed so]'') and disagreements about relevance. This is related to the Gricean maxim of relation where it suggests all information should be relevant to the discourse. ``Definitions and clarity'' (P2) includes arguments that attempt to clarify terms, draw distinctions, or critique the definitions already in place (\emph{e.g.}, in the phrase ``trivially true''). Fragments such as ``ambiguous'' and ``vague term'' are directly linked to the Gricean maxim of manner (clarity) that discourages ambiguity in conversations.

A final pattern, ``personal and anecdotal'', is commonly associated with evidence provided from a first-person perspective. It includes phrases about what things ``sound like'' or ``look like'', for example, as well as evidence from personal experience. While the ``relevance and presumption'' pattern contains ``never said'', for example, suggesting that an assumption has been (incorrectly) made, this pattern contains ``never felt'', as in ``I never felt that X''. Consistent with this interpretation, this final pattern is the most loaded on the personal pronoun ``I''.

Table~\ref{linkage_table} also shows the relative frequency with which an argument pattern appears in a comment in both r/CMV and LessWrong. (We restrict our consideration to comments that have at least one hit in one of the argument patterns, and we eliminate the meta topic.) The two corpora show broad similarities, with ``relevance and presumption'' being the least common pattern and ``personal and anecdotal'' the most common.

\begin{table*}[]
    \centering
    \begin{tabular}{l|l|c|c}
    Pattern & Sample Bigrams & r/CMV & LessWrong \\ \hline
    1. Relevance \& Presumption & completely irrelevant, fundamentally wrong, never said & 14.4\% & 7.4\% \\
    2. Definitions \& Clarity & defined as, distinction between, clarifying question, don't see & 12.8\% & 15.0\% \\
    3. Deduction \& Certainty & can't prove, objectively true, doesn't change, an opinion & 11.2\% & 18.1\% \\
    4. Causation \& Examples & directly correlated, mainly because, certain contexts & 15.8\% & 15.8\% \\
    5. Induction \& Probability & quite possible, more likely, almost certainly, average person & 16.1\% & 20.7\% \\
    6. Personal \& Anecdotal & seemed like, personal anecdote, never felt, definitely agree & 29.6\% & 23.0\% \\
    \end{tabular}
    \caption{Automatically-extracted argument patterns. Our method finds six clusters, corresponding to distinct argument-making pragmatics such as questioning relevance, discussing definitions, and deductive and certainty-based reasoning. The detected patterns appear at different rates, although in roughly similar rank-order in the two corpora, r/ChangeMyView and LessWrong.}
    \label{linkage_table}
\end{table*}

\begin{table*}
    \begin{tabular}{p{3.25in}|p{3.25in}}
     {\bf 1. Relevance and Presumption} & {\bf 2. Definitions and Clarity} \\ \hline
     Murder \textcolor{red}{has nothing} to do with trade.
    Rape \textcolor{red}{has nothing} to do with trade.
    Smoking weed \textcolor{red}{has nothing} to do with trade.
    Speeding \textcolor{red}{has nothing} to do with trade. [...]
    Each and every one of them is a person doing something against the rules of society. \vspace{-0.2cm} \newline \hphantom{} \hspace{-0.3cm} \hrulefill  \vspace{-0.1cm} \newline 
    It's not \textcolor{red}{supposed to} prevent collusion, that's on the FEC to enforce.  Again, if collusion is happening, that is illegal and should be prosecuted. That \textcolor{red}{says nothing} about whether Citizens United was the right decision or not. \vspace{-0.2cm} \newline \hphantom{} \hspace{-0.3cm} \hrulefill  \vspace{-0.1cm} \newline
    I \textcolor{red}{never said} that nobody else can enjoy it. [...] A straight person can go to a pride festival because they support it and there is \textcolor{red}{absolutely} \textcolor{red}{nothing wrong} with that. 
    & I think it's more valuable to \textcolor{red}{specifically} and \textcolor{red}{explicitly} distinguish between ``racism" and ``systemic racism" at least in common parlance. [...] If you think of it like a Venn diagram, this overlap between the academic and colloquial definitions just \textcolor{yellow}{leads to} \textcolor{red}{ambiguity} in meaning. \vspace{-0.2cm} \newline \hphantom{} \hspace{-0.3cm} \hrulefill  \vspace{-0.1cm} \newline  I'm \textcolor{red}{not sure} I follow the point you’re \textcolor{red}{trying to} make. People are \textcolor{red}{obviously} equipped to apply reason to problems---we do so every day, so we are \textcolor{red}{clearly} equipped to do it. We are simply trained not to apply that reason to \textcolor{yellow}{certain} categories of problems like overconsumption. [...] We have the mental equipment, we choose not to use it. That's \textcolor{red}{distinctly different} from not being ``mentally equipped'' to do it. \\ \hline
     {\bf 3. Deduction and Certainty} & {\bf 4. Causation and Examples} \\ \hline
     
     If you do accept that all the \textcolor{red}{facts} were reported on (maybe with bias), then the \textcolor{red}{question becomes} not whether there was information withheld but [...] \vspace{-0.2cm} \newline \hphantom{} \hspace{-0.3cm} \hrulefill  \vspace{-0.1cm} \newline 
     If I can only be sure of my own perceptions (which is \textcolor{red}{an assumption} in and of itself) and all other \textcolor{red}{evidence} is equally invalid, the pursuit of any knowledge becomes pointless. \vspace{-0.2cm} \newline \hphantom{} \hspace{-0.3cm} \hrulefill  \vspace{-0.1cm} \newline  There is different [sic] between claiming something and offering \textcolor{red}{no proof} and claiming something and \textcolor{red}{proving} the opposite is wrong.
  & 
     That said, different personalities \textcolor{red}{react differently} to the same things. A person who complains about something \textcolor{red}{isn't necessarily} weaker or more sensitive or less \textcolor{red}{confident} in their identity: they are \textcolor{red}{perhaps} more disagreeable and more assertive. \vspace{-0.2cm} \newline \hphantom{} \hspace{-0.3cm} \hrulefill  \vspace{-0.1cm} \newline 
    More broadly, assessing whether something is a ``disability" \textcolor{red}{really depends} on whether it is impacting the particular task domain being \textcolor{red}{considered} [...] how an individual is affected by their autism \textcolor{red}{can vary} \textcolor{yellow}{a lot} \textcolor{red}{depending on} the individual. \\ \hline
     {\bf 5. Induction and Probability} &  {\bf 6. Personal and Anecdotal} \\ \hline
     Males are at the forefront of a big swath of social dysfunction. If a woman has been beaten or killed, she was \textcolor{red}{most likely} beaten or killed by a man. [...] Men are \textcolor{red}{more likely} to die of suicide. [...] 
\vspace{-0.2cm} \newline \hphantom{} \hspace{-0.3cm} \hrulefill  \vspace{-0.1cm} \newline  It is very \textcolor{red}{possible} that if we achieved a technological singularity [...] the rate at which we can make new discoveries would accelerate so much that 1000 years of medical research \textcolor{red}{might be} accomplished in 10 years.
Is it \textcolor{red}{unlikely} that this will happen? \textcolor{red}{Possibly}. Is it a greater than 0\% \textcolor{red}{possibility} that this will happen? \textcolor{yellow}{Absolutely}.
 \vspace{-0.2cm} \newline \hphantom{} \hspace{-0.3cm} \hrulefill  \vspace{-0.1cm} \newline  The main issue with \textcolor{yellow}{any form} of vigilantism is that there is \textcolor{yellow}{absolutely} \textcolor{red}{no guarantee} or \textcolor{red}{likelihood} that the vigilantes would be any better than the people they replace.

     & Also, kids learn \textcolor{red}{a lot} through physical interaction. I \textcolor{red}{remember} growing up and playing on merry go rounds and teeter totters. Those gave me, and a ton of other kids, an intuitive understanding of how fulcrums and centripetal force works. We \textcolor{red}{didn't know} what they were called but we knew intuitively that the merry go round would \textcolor{red}{try} to throw us off the faster it went [...] \vspace{-0.2cm} \newline \hphantom{} \hspace{-0.3cm} \hrulefill  \vspace{-0.1cm} \newline 
This one \textcolor{red}{feels} far more important to me. People have \textcolor{red}{never been} so interconnected. Information has \textcolor{red}{never} spread as far or as quickly as it does today. This has also \textcolor{yellow}{led} to a decline in other forms of news. \vspace{-0.2cm} \newline \hphantom{} \hspace{-0.3cm} \hrulefill  \vspace{-0.1cm} \newline 
It turns out that it's \textcolor{red}{kind of} challenging to describe in very \textcolor{yellow}{explicit} terms, but I'll \textcolor{red}{try} to at least give a reasonable broad overview of how I personally \textcolor{red}{felt}. 
    \end{tabular}
    \caption{Examples of the classification in action, using excerpts from comments in r/ChangeMyView. Each comment is dominated by a particular pattern; red words and bigrams are from the dominant pattern, yellow from other patterns. All of these comments received a ``delta'', indicating that the comment changed at least one reader's point of view.\label{examples}}
\end{table*}


\subsection{Argument Efficacy}

Table~\ref{cmv_delta} shows the ``$\Delta$ bonus'', or the relative likelihood that a comment dominated by one of the six argument patterns is tagged as changing a person's point of view in the r/CMV data. Immediately apparent are the large differences in outcome for the different patterns. One of the least successful pattern, ``relevance and presumption'' is 42\% less likely to lead to a (reported) change in someone's point of view, compared to the most successful pattern, ``causation and examples''.

These effects persist in both size and direction even when controlling for argument semantics. We can see that (for example) the negative effect associated with questioning relevance persists across discussions as varied as gun control, moral duties to animals and children, race, religion and culture, and sex and gender. (The ``induction and probability'' pattern has the most variability and, in discussions of both race and morality, the usually positive effect disappears, possibly reflecting an aversion to the use of stereotypes about ``likely'' characteristics.) Table~\ref{scores} shows that these patterns extrapolate, to a lesser extent, when considering popularity through upvote/downvote rating.

\subsection{A Diversity of Preferences}

Our final analysis considers the diversity of preferences at the individual level. Just as there are different types of arguments, we ask, are there different types of argument-makers?

For simplicity, and so that each individual's preferences are well-sampled, we restrict to users with at least twenty comments, and calculate each user's average distribution over the six argument patterns. A simple PCA analysis (Table~\ref{factor}) then reveals two main components to user preferences in argument-making. The first component, which explains 67\% of the variance, corresponds to a ``personal--impersonal'' axis; users high on this first component tend to prefer arguments that draw on the ``personal and anecdotal'' pattern.

The second component explains 13\% of the variance, and corresponds to a preference for both ``causation and examples'' and ``induction and probability''. We refer to this as the ``concrete--abstract'' axis; users high on this axis prefer reasoning about causes, examples, and relative probabilities as opposed to logical certainties, relevance, or definitions.

Most notably, while successful arguments sometimes have a personal aspect (Table~\ref{cmv_delta}), the most successful argument-makers are found in the impersonal-concrete quadrant ($+32\%$ $\Delta$ bonus, compared to $-28\%$ for personal-abstract). The apparent contradiction arises from the fact that we have focused on the argument-makers who achieve repeated success. Personal arguments can often work, but users who are reliably successful often take a more impersonal (and concrete) stance.

\section{Discussion}

Traditional accounts of argument-making have focused on the inductive-deductive distinction. Our results suggest that, when it comes to pragmatics, psychologically distinct modes of persuasion emerge that go beyond the inductive-deductive distinction, with distinct properties and rates of real-world success. We don't just persuade each other by arguing probabilities. We also try to clarify our definitions, present hypotheticals, speak from personal experience, and make our---or our opponent's---assumptions explicit. This highlights a key, and often-neglected issue in the study of argument making: we argue with another, usually specific, person, and---contrary to normative results such as Aumann's Agreement theorem~\cite{hanson}---our disagreements may not always be about the information to hand, but how, for example, we divide up the space of possibilities.

Remarkably, our findings suggest that argument patterns preserve their appeal across different contexts. Questioning relevance is just as unpopular in discussions about sexuality as it is in talk about the politics of crime. First-person testimony---including talk about what one has heard, feels is true, or has personally experienced---is enduringly popular across both communities and in domains, such as physics or AI, where it might seem beside the point.

Furthermore, our results provide new insight into the diversity of argument preferences, about which little is currently known~\cite{knauff2021handbook}. Rather than dividing the world into (say) logicians and probabilists, we find a dominant role for preferences along a personal-impersonal axis. The shifting influence of individuals at different points along this axis may help explain recent results in cultural evolution~\cite{Scheffere2107848118}, that report a large-scale shift in discourse from impersonal rationality to a more intuitive and first-personal style.

Finally, the fact that our pattern lexicon emerges from the simple LIWC categories of ``tentative'' and ``certain'' shows how common terms can signal distinct, and distinctly-successful forms of argument-making. While unsupervised techniques such as topic modelling have been successful in clustering documents with semantic similarity, potentially pragmatic phrases and patterns are often discarded as stopwords or ``junk'' topics. Simple information-theoretic methods, however, suggests there is far more than meets the eye.


\hspace{-0.3cm} \hrulefill 
\begin{table}[H]
    \centering
    \begin{tabular}{l|c|c}
    & Personal-Impersonal & Concrete-Abstract  \\ \hline
     Relevance & $-0.20$  & $-0.67$ \\
     Definitions & $-0.25$ & $-0.25$ \\
     Deduction & $-0.42$ & $-0.49$ \\
     Causation & $-0.43$ & $\mathbf{+0.51}$ \\
     Induction & $-0.29$ & $\mathbf{+0.16}$ \\
     Personal & $\mathbf{+0.59}$ & $-0.24$ 
    \end{tabular}
    \caption{Primary factors in argument-making preferences; factor loadings from a PCA analysis of the 22,493 participants from r/CMV with at least twenty comments.}
    \label{factor}
\end{table}

\begin{table*}
\begin{tabular}{l|l|l|l|l|l|l|l}
(r/CMV topic) & P($\Delta$) (\%) & Relevance & Definitions & Deduction & Causation & Induction & Personal \\ \hline
[all] &  $1.35$ &  $\mathbf{-19.4\%^{\star\star\star}}$ &  $\mathbf{-18.5\%^{\star\star\star}}$ &  $\mathbf{-20.2\%^{\star\star\star}}$ & $\mathbf{+23.0\%^{\star\star\star}}$ &     $\mathbf{+15.4\%^{\star\star\star}}$ &   $\mathbf{+19.7\%^{\star\star\star}}$ \\ \hline
black-white-culture &  $1.04$ &  $\mathbf{-24.6\%^{\star\star\star}}$ &  $\textcolor{gray}{-9.9\%}$ &  $\mathbf{-21.2\%^{\star\star\star}}$ &  $\mathbf{+35.2\%^{\star\star\star}}$ &      $\textcolor{gray}{+5.0\%}$ &   $\mathbf{+15.5\%^{\star\star\star}}$ \\
man-sex-child &  $1.18$ &  $\mathbf{-21.8\%^{\star\star\star}}$ &  $\mathbf{-13.3\%^{\star\star\star}}$ &  $\mathbf{-27.4\%^{\star\star\star}}$ &  $\mathbf{+28.2\%^{\star\star\star}}$ &    $\mathbf{+9.7\%^{\star\star}}$ &   $\mathbf{+24.7\%^{\star\star\star}}$ \\
country-war-power &  $1.26$ &  $\mathbf{-15.2\%^{\star\star\star}}$ &  $\mathbf{-17.1\%^{\star\star\star}}$ &  $\mathbf{-23.7\%^{\star\star\star}}$ &  $\mathbf{+21.9\%^{\star\star\star}}$ &   $\mathbf{+19.7\%^{\star\star\star}}$ &   $\mathbf{+14.5\%^{\star\star\star}}$ \\
god-music-art &  $1.74$ &  $\mathbf{-14.5\%^{\star\star\star}}$ &    $\mathbf{-17.7\%^{\star\star\star}}$ &  $\mathbf{-24.5\%^{\star\star\star}}$ &  $\mathbf{+20.8\%^{\star\star\star}}$ &      $\mathbf{+9.3\%^{\star\star\star}}$ &  $\mathbf{+26.6\%^{\star\star\star}}$ \\
food-animals-eat &  $1.70$ &  $\mathbf{-15.7\%^{\star\star\star}}$ &  $\mathbf{-19.2\%^{\star\star\star}}$ &  $\mathbf{-22.9\%^{\star\star\star}}$ &  $\mathbf{+22.3\%^{\star\star\star}}$ &  $\mathbf{+18.8\%^{\star\star\star}}$ &  $\mathbf{+16.8\%^{\star\star\star}}$ \\
law-rape-legal &  $1.21$ &  $\mathbf{-24.2\%^{\star\star\star}}$ &  $\mathbf{-16.9\%^{\star\star\star}}$ &  $\mathbf{-13.9\%^{\star\star\star}}$ &  $\mathbf{+20.2\%^{\star\star\star}}$ &   $\mathbf{+20.4\%^{\star\star\star}}$ &  $\mathbf{+14.4\%^{\star\star\star}}$ \\
human-moral-exist & $1.19$ &  $\mathbf{-14.5\%^{\star\star}}$ &  $\mathbf{-20.3\%^{\star\star\star}}$ &  $\textcolor{gray}{-1.1\%}$ &  $\mathbf{+18.5\%^{\star\star\star}}$ &   $\textcolor{gray}{+8.5\%}$ &  $\mathbf{+8.9\%^{\star}}$ \\
pay-job-vote & $1.26$ &  $\mathbf{-20.6\%^{\star\star\star}}$ &    $\mathbf{-19.4\%^{\star\star\star}}$ &    $\mathbf{-19.9\%^{\star\star}}$ &  $\mathbf{+27.2\%^{\star\star\star}}$ &      $\mathbf{+20.9\%^{\star\star\star}}$ &   $\mathbf{+11.7\%^{\star\star\star}}$ \\
school-education-high &  $1.61$ &  $\mathbf{-20.2\%^{\star\star\star}}$ &  $\mathbf{-24.8\%^{\star\star\star}}$ &  $\mathbf{-12.8\%^{\star\star}}$ &  $\mathbf{+19.2\%^{\star\star\star}}$ &  $\mathbf{+13.1\%^{\star\star}}$ &  $\mathbf{+25.4\%^{\star\star\star}}$ \\
\end{tabular}
\caption{The $\Delta$ bonus for different argument patterns, as a function of semantic context, in r/CMV. $\Delta$ bounus is the percentage increase (or decrease) in probability that a comment received a ``changed my view'' tag in r/CMV. In each case we show the percentage increase (or decrease) in probability that a comment received a ``changed my view'' tag. Semantics are labelled by the top three words in the topic model and correspond to intuitive themes (\emph{e.g.}, posts in the ``gun-crime-violence'' topic include arguments about crime and gun control.) Error bars are small (less than $\pm 0.1$ percentage points in most cases) and are not shown for clarity. $p$-value labels are for differences between outcomes controlling for semantics (\emph{e.g.}, ``relevance'' performs significantly worse in discussions under the gun-crime-violence topic, compared to the average outcome for the six patterns in that topic.). $\star$ indicates $p<0.05$; $\star\star$, $p<0.01$; $\star\star\star$, $p<10^{-3}$. \label{cmv_delta}}
\end{table*}

\begin{table*}
\begin{tabular}{c}

\begin{tabular}{p{3.25cm}|l|l|l|l|l|l|l}
 (r/CMV) & Average Score & Relevance & Definitions & Deduction & Causality & Induction & Personal \\ \hline
all &  $2.84$ &  $\mathbf{-0.10^{\star\star\star}}$ &  $\mathbf{-0.14^{\star\star\star}}$ &  $\mathbf{-0.24^{\star\star\star}}$ &  $\mathbf{+0.19^{\star\star\star}}$ &    $\textcolor{gray}{-0.01}$ &  $\mathbf{+0.29^{\star\star\star}}$ \\ \hline
black-white-culture &  $2.70$ &   $\mathbf{-0.18^{\star\star}}$ &  
$\mathbf{-0.19^{\star\star\star}}$ &   $\mathbf{-0.21^{\star\star}}$ &    $\mathbf{+0.22^{\star\star}}$ &    $\textcolor{gray}{+0.01}$ &  $\mathbf{+0.25^{\star\star}}$ \\
man-sex-child &  $3.08$ &  $\mathbf{-0.19^{\star\star\star}}$ &   $\mathbf{-0.15^{\star\star\star}}$ &  
$\mathbf{-0.29^{\star\star\star}}$ &
$\mathbf{+0.27^{\star\star\star}}$ &
$\textcolor{gray}{0.0}$ &  
$\mathbf{+0.36^{\star\star\star}}$ \\
country-war-power &  $2.91$ &  
$\textcolor{gray}{-0.07}$ &
$\mathbf{-0.13^{\star}}$ &  $\mathbf{-0.31^{\star\star\star}}$ & 
$\mathbf{+0.18^{\star}}$ &    $\textcolor{gray}{+0.03}$ &  $\mathbf{+0.3^{\star\star\star}}$ \\
god-music-art &  $3.12$ &  
$\textcolor{gray}{-0.03}$ &  
$\mathbf{-0.13^{\star}}$ &
$\mathbf{-0.32^{\star\star\star}}$ &
$\textcolor{gray}{-0.08}$ &
$\textcolor{gray}{+0.02}$ &  
$\mathbf{+0.41^{\star\star\star}}$ \\
food-animals-eat &  $3.00$ &  
$\textcolor{gray}{-0.05}$ &
$\textcolor{gray}{-0.11}$ & 
$\mathbf{-0.35^{\star\star\star}}$ &
$\mathbf{+0.21^{\star\star}}$ &    $\textcolor{gray}{-0.08}$ &   $\mathbf{+0.38^{\star\star\star}}$ \\
law-rape-legal  &  $2.69$ &  
$\textcolor{gray}{-0.1}$ &     $\mathbf{-0.19^{\star\star\star}}$ &    $\textcolor{gray}{-0.04}$ &  
$\mathbf{+0.11^{\star}}$ &    $\textcolor{gray}{+0.05}$ &  $\mathbf{+0.17^{\star\star\star}}$ \\
human-moral-exist & $2.17$ &  
$\textcolor{gray}{+0.01}$ & 
$\mathbf{-0.12^{\star\star}}$ &  
$\mathbf{-0.13^{\star\star}}$ &  
$\mathbf{+0.15^{\star\star}}$ &     $\textcolor{gray}{-0.01}$ &  $\mathbf{+0.1^{\star}}$ \\
pay-job-vote & $2.66$ &   
$\mathbf{-0.08^{\star}}$ &     $\textcolor{gray}{-0.08}$ &  $\mathbf{-0.23^{\star\star\star}}$ &
$\mathbf{+0.22^{\star\star\star}}$ &
$\textcolor{gray}{+0.02}$ &
$\mathbf{+0.15^{\star\star\star}}$ \\
school-education-high &  $2.85$ &    
$\mathbf{-0.16^{\star}}$ &
$\mathbf{-0.18^{\star\star\star}}$ &
$\textcolor{gray}{-0.03}$ &
$\mathbf{+0.22^{\star\star}}$ &    $\textcolor{gray}{-0.01}$ &  $\mathbf{+0.16^{\star\star}}$ \\ 
\end{tabular} \\ \\
\begin{tabular}{p{3.25cm}|l|l|l|l|l|l|l}
(LessWrong) & Average Score & Relevance & Definitions & Deduction & Causality & Induction & Personal \\ \hline
all & $3.53$ &
$\textcolor{gray}{+0.02}$ & $\mathbf{-0.11^{\star\star\star}}$ &
$\mathbf{-0.17^{\star\star\star}}$ & $\mathbf{+0.09^{\star\star}}$ & $\mathbf{-0.37^{\star\star\star}}$ & $\mathbf{+0.54^{\star\star\star}}$\\ \hline
utility-function-agent & $2.36$ & $\textcolor{gray}{-0.07}$ &
$\textcolor{gray}{-0.05}$ &
$\textcolor{gray}{+0.03}$ &
$\textcolor{gray}{-0.08}$ & $\textcolor{gray}{-0.06}$
& $\mathbf{+0.23^{\star\star\star}}$ \\
universe-brain-physics & $2.89$ & $\textcolor{gray}{+0.2}$&
$\mathbf{-0.14^{\star}}$ &
$\textcolor{gray}{+0.04}$ & $\textcolor{gray}{-0.06}$ & $\mathbf{-0.32^{\star\star\star}}$ & $\mathbf{+0.29^{\star\star\star}}$ \\
money-years-risk & $3.81$ &
$\textcolor{gray}{-0.13}$ &
$\textcolor{gray}{-0.04}$ & $\mathbf{+0.16^{\star\star}}$ & $\textcolor{gray}{-0.01}$ & $\mathbf{-0.16^{\star\star}}$ & $\mathbf{+0.19^{\star\star\star}}$ \\
person-moral-bad & $4.23$ &
$\textcolor{gray}{+0.1}$ &
$\mathbf{-0.19^{\star\star\star}}$ &
$\mathbf{-0.22^{\star\star\star}}$ &
$\textcolor{gray}{0.0}$ & $\mathbf{-0.3^{\star\star\star}}$ & $\mathbf{+0.61^{\star\star\star}}$ \\
ai-humans-intelligence & $3.13$ & $\textcolor{gray}{-0.15}$ &
$\textcolor{gray}{0.0}$ &
$\textcolor{gray}{-0.04}$ &
$\textcolor{gray}{-0.12}$ &
$\textcolor{gray}{-0.01}$ &
$\mathbf{+0.32^{\star\star\star}}$ \\

\end{tabular} \\ \\

\end{tabular}

\caption{LessWrong and r/CMV score bonus for different argument pragmatics, as a function of semantic context. In each case we show the change in the number of upvotes on comments, conditional upon semantics, associated with domination by one argument pattern over another.  Paralleling the case of the r/CMV $\Delta$-bonus result, we see large and significant differences in the overall rating of posts containing different argument patterns, which are largely stable in direction across semantic contexts. ``Personal and Anecdotal'' is strongly preferred in every context. While the argument patterns that convince tend to be the ones that others upvote, we see occasional divergences between upvoting and view-changing: for example, inductive and probabilistic arguments are more likely to receive a delta compared to baseline, but are not significantly upvoted. Differences also emerge between r/CMV and LessWrong, most notably in the latter's bias against inductive arguments, and tolerance of debates over relevance.\label{scores}}
\end{table*}

\clearpage
\section{Acknowledgements}

We thank Paul Smaldino, Ben Hoffmann, and our three anonymous referees for helpful feedback. R.W.N.\ and S.D.\ were supported in part by the National Science Foundation under Grant No.\ SES-1948887, and by the Survival and Flourishing Fund.

\bibliographystyle{apacite}

\setlength{\bibleftmargin}{.125in}
\setlength{\bibindent}{-\bibleftmargin}

\bibliography{main}

\end{document}